%% file: example.tex
\documentclass{article}

\usepackage[final]{corl_2026} 

\usepackage{graphicx}

\title{Multi-Objective Adaptation After Deployment: A Position on World Models for Remote Robot Teams}

\author{
  Jamie Santos and Kagan Tumer\\
  Collaborative Robotics and Intelligent Systems (CoRIS) Institute\\
  Oregon State University \\
  \texttt{\{santjami,Kagan.Tumer\}@oregonstate.edu} \\
}

\usepackage{glossaries}
\makeglossaries
\input{glossary}

\begin{document}
\maketitle


\begin{abstract}
Autonomous robot teams deployed in remote, unstructured environments may face changes in the environment, hardware, and team composition that degrade previously effective policies. Shifting mission objectives can be handled by selecting a different solution from an existing Pareto set, but shifting deployment conditions can reorder the set itself, making previously dominated policies competitive or new policies necessary. However, evaluating these policies in new contexts while minimizing risk to the hardware remains an open problem. \textit{We argue that, although largely untested in this domain, world models merit investigation for post-deployment multi-objective adaptation.} By predicting policy outcomes under updated estimates of current conditions, world models could enable teams to re-evaluate existing policies without physically executing each candidate, provided predicted performance changes reflect those realized on deployed systems. We identify research priorities for model-based policy re-evaluation, validation of predicted against realized performance, and incorporating shared information into world models to support coordinated team adaptation.
\end{abstract}



\section{Introduction}
Autonomous robot teams are increasingly important to sustaining long-term missions in which human intervention is difficult, such as space and marine environments. Multi-robot coordination can improve efficiency and fault tolerance in such missions \citep{parker2002alliance}, and efforts such as NASA's CADRE are investigating how teams of rovers can coordinate with minimal human intervention \citep{de2024multi}. Single-robot systems such as the Curiosity Mars rover or MBARI's Tethys Autonomous Underwater Vehicle \citep{vasavada2022mission,bellingham2025marine} have also already demonstrated autonomous navigation and task-planning, and autonomy stacks such as NASA's NeBula aim to integrate these capabilities \citep{agha2022nebula}. However, sustaining autonomy over extended deployments remains challenging. Environmental conditions, hardware performance, team composition, and mission objectives can change in ways that undermine previously effective policies and affect team coordination and overall mission performance  \citep{kunze2018artificial}. 

These challenges are particularly acute in unstructured environments, where conditions may shift unpredictably. Sandstorms may coat solar panels with dust, hardware components may degrade or become damaged, and team members may become incapacitated \citep{vasavada2022mission, hawes2017strands}. Such changes can invalidate assumptions made during training and shift the performance dominance ranking of available policies. A policy that was previously dominated may become competitive under new conditions, while previously effective policies may no longer achieve their expected performance. For example, when the front right drive wheel actuator of the Spirit Mars rover failed, operators began driving the rover backwards from then on \citep{callas2015mars, webster_2009}. The locked wheel also became an unexpected scientific tool; its trenching effect produced the discovery of opaline silica deposits, which may indicate past locally habitable conditions \citep{callas2015mars}. This example illustrates how hardware changes can alter the utility of available behaviors. Whereas changing mission preferences may require selecting a different solution from an existing Pareto set, changes in deployment conditions may alter the relative performance of stored policies, making previously dominated solutions competitive or requiring new policies when existing alternatives are insufficient. Yet, adapting policies in situ is difficult because physical experimentation can be costly or risky, particularly when robots operate far from human operators or have limited computational and energy resources.

World models are a candidate for post-deployment adaptation because they allow a robot to learn from deployment experience and use imagined rollouts to re-evaluate existing policies \citep{hafner2019dream} under recent conditions \citep{domberg2026self}, without physical trial and error. Recent work by Domberg et al. \cite{domberg2026self} has demonstrated the ability to adapt world models by tracking model prediction residuals and finetuning the models when the context is out-of-distribution. This could help teams identify suitable existing policies, including alternatives that were previously dominated, and determine when new policies are needed, reducing reliance on physical experimentation. However, the approach depends on whether decentralized policy adaptations can preserve effective team coordination. We argue that world models merit further investigation as tools for continually re-evaluating and adapting multi-objective team policies during deployment, with particular attention to the reliability of predicted improvements and their effects on team-level behavior.

\begin{figure}
    \centering
    \includegraphics[width=1\linewidth]{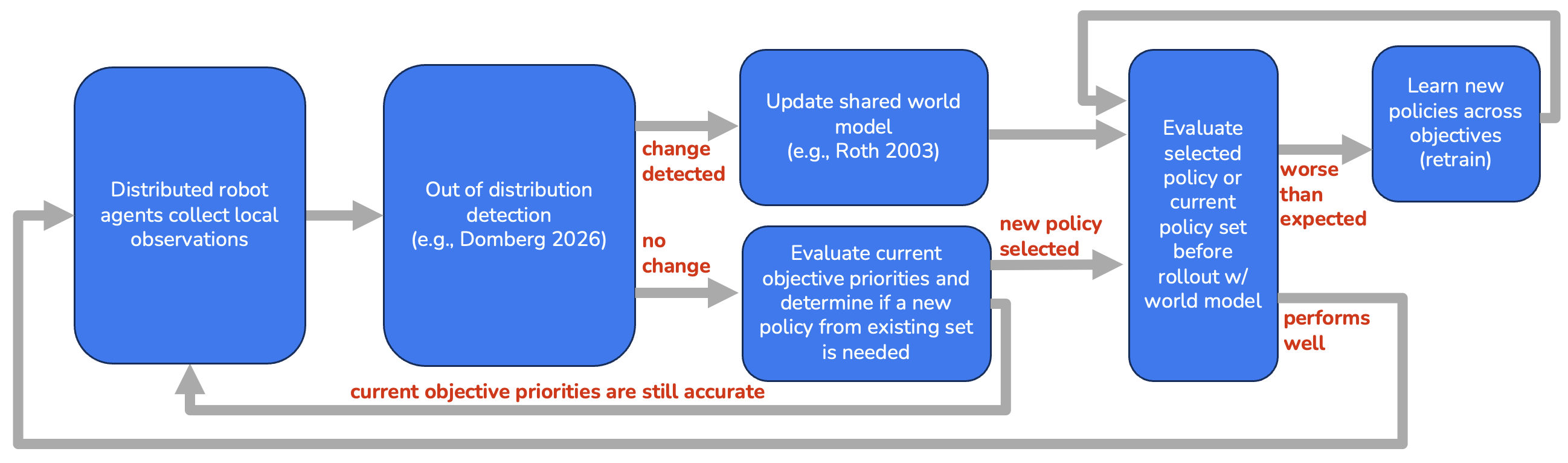}
    \caption{This is a proposed pipeline for maintaining a shared, current representation of the world using world models after deployment. Distributed agents collect local observations, model predictions are tracked against realized performance, and if the context is out of distribution, the model can be updated. The model is used to evaluate policies when objectives change, or possibly even learn new policies with the updated world models.}
    \label{fig:placeholder}
\end{figure}

\section{Background: World Models and Existing Adaptation Approaches}

\subsection{World Models for Post-Deployment Learning}
Preparing robots for deployment can improve their ability to handle uncertainty, but cannot eliminate the need to adapt to unforeseen conditions. Domain randomization can improve sim-to-real transfer by exposing policies to a distribution of environmental conditions during training \citep{radosavovic2024real} while simulation during pretraining and methods that combine offline demonstrations with limited online adaptation can reduce reliance on real-world experience \citep{li2025sim, makoviychuk2021isaac, brohan2022rt}. However, these approaches cannot guarantee effective adaptation to deployment conditions that are inadequately represented during training or modeling. This challenge is especially pertinent to long-term deployments in unstructured environments such as Mars or Antarctica.

Post-deployment adaptation is particularly challenging for remote robots, where opportunities for safe physical experimentation may be limited. Current systems in high-stakes regions such as the Moon rely on heavy human intervention to maintain mission safety, but this approach becomes less manageable as the number of agents increases and robots are used to take on more diverse roles. Testing unverified policies in environments where human intervention is difficult can have serious consequences; unexpected behavior may render a robot inoperable. This constraint in particular motivates approaches that can evaluate candidate behaviors using recent deployment experience while reducing reliance on costly or risky physical experimentation.

World models offer a potential means of supporting this process. By learning a compact predictive representation of environmental dynamics from onboard data, they allow agents to evaluate candidate behaviors through imagined interactions rather than relying exclusively on physical experimentation \citep{hafner2019dream}. DayDreamer, for example, demonstrated that world-model-based reinforcement learning can train and adapt policies on physical robots with substantially less interaction than conventional approaches \citep{wu2023daydreamer}. These results suggest that learned models could provide a useful basis for evaluating and adapting robot behaviors after deployment. A key challenge, however, is maintaining model accuracy as deployment conditions change. Domberg et al. \citep{domberg2026self}, as aforementioned, demonstrate one approach to addressing this challenge by monitoring prediction residuals and triggering online model updates when residuals deviate substantially from their expected values. However, extending such approaches to multi-objective robot teams requires accounting for changes not only in environmental conditions but also in team capabilities and mission requirements.
\vspace{-5pt}
\subsection{Approaches to Maintaining and Adapting Policy Repertoires}
Several approaches address the challenge of maintaining effective behaviors across different tasks and conditions. Multi-task learning trains policies across distributions of tasks or objectives \citep{wang2021bridging}, while quality-diversity methods such as MAP-Elites generate repertoires of diverse policies \citep{mouret2015illuminating}, including policies optimized across multiple objectives \citep{pierrot2022multi}. These approaches can provide a range of behaviors from which a robot or team can select as conditions change. However, they do not guarantee that available policies will remain effective under unforeseen deployment conditions.

Meta-learning methods such as Model-Agnostic Meta-Learning (MAML) instead aim to enable rapid adaptation from limited experience \citep{gharoun2024meta, finn2017model}, with extensions such as \gls{difmaml} addressing adaptation in multiagent settings without requiring centralized coordination \citep{kayaalp2022dif}. However, these methods still depend on informative experience, which may be costly or risky to obtain during deployment. 

World models offer a complementary approach to these methods. Rather than relying exclusively on updating individual policies from new experience, a learned dynamics model can be updated using deployment data collected under different policies and then used to re-evaluate multiple existing policies. In a multi-objective setting, this could allow a team to assess its policy repertoire across competing objectives as conditions change, potentially identifying suitable existing policies before committing resources to training new ones. Whether model-based evaluations can reliably support these decisions under changing deployment conditions remains an open question.

\vspace{-5pt}
\subsection{Reliability and Coordination Challenges}
For multiagent systems, maintaining reliable world models introduces additional challenges. Partial observability and communication constraints may limit the information available to individual agents, making it difficult to construct accurate representations of joint team dynamics. Changes in team composition or individual robot capabilities may further complicate predictions of team behavior. Even when individual agents identify promising policy updates, independently adapting agents may not produce improved team-level performance. In contrast, early work in multiagent world models demonstrates the efficacy of maintaining both individual and team-level world models on each agent in scenarios with high-latency communication \citep{1249244}. Recent work has begun to investigate world models for multiagent task learning and planning \citep{zhao2026learning,zou2026ma}, but it remains unclear whether these models can support reliable policy improvements under unforeseen deployment changes while preserving effective team coordination and acceptable tradeoffs across objectives.

\section{World Models as Tools for Adaptation}
Over long-term deployments, changes in terrain, hardware, team composition, or mission objectives may exceed the adaptability of existing methods \citep{kunze2018artificial}. A robot team originally deployed to conduct scientific experiments, for instance, may eventually be reassigned to perform logistical or maintenance tasks, so adaptation must cover not only low-level behaviors but also higher-level decisions about which policies to deploy. Post-deployment adaptation must therefore account for both changing mission priorities and changes in the behaviors available to meet them.

Furthermore, determining \textit{when} adaptation is necessary is itself a challenge. A change in observed performance may indicate that a policy has become less effective under current conditions, but becoming less effective for one objective may also correspond to increased performance in another; for example, a rover that begins to cover less ground due to hardware degradation may also make more scientific discoveries. Environmental changes and hardware failures may also alter the tradeoffs the team as a whole can achieve, potentially rendering previously discovered Pareto-optimal policies now suboptimal. Distinguishing between a degraded policy, a shifted trade-off, and a change in priorities requires evaluating policy performance in the context of both current conditions and competing objectives. Whereas changes in deployment conditions may require updating the model and re-evaluating the available policy repertoire, changes in mission priorities may only require selecting a different policy from the existing Pareto set.

World models could support this process by enabling ongoing re-evaluation of policy repertoires during deployment. Figure \ref{fig:placeholder} demonstrates a possible pipeline to accomplish this in a multiagent system. Distributed agents can collect local observations that can be used to update a shared world model (either updated via a distributed or centralized manner depending on the system architecture) \citep{1249244}. By combining experiences collected across agents, a shared model could capture team-level interactions and changes that may not be observable from any individual agent's local experience alone. Differences between model predictions and observed outcomes can indicate that the model is no longer accurately representing current deployment conditions. Such discrepancies could trigger model updates, followed by re-evaluation of existing policies using the updated model. Policy updates or the generation of new policies would then be considered if the existing repertoire no longer provides suitable alternatives \citep{domberg2026self}. By updating a learned dynamics model with newly collected experience, potentially while giving greater weight to recent observations, a robot team could maintain predictions that better reflect its current operating conditions. The model could then be used to re-evaluate existing policies, identify which remain suitable, and determine when further policy adaptation may be necessary. This approach could reduce reliance on physical experimentation, but its effectiveness depends on whether sufficiently accurate models can be learned and updated from limited deployment experience. A reliable framework must therefore assess not only predicted policy performance, but also whether predicted improvements transfer to the deployed system. 

\section{Research Priorities and Conclusion}
Realizing the potential of world models for multi-objective post-deployment adaptation requires evaluating both the reliability of model predictions and the effectiveness of the resulting policy updates. First, research should investigate how deployed systems can detect \textit{when} adaptation is necessary, distinguishing changes in environmental conditions from changes in mission priorities. Second, model-based adaptation should be evaluated by comparing predicted and realized improvements across multiple objectives and under changes in hardware, environment, and team composition. Finally, evaluations should consider whether local policy updates preserve team-level coordination and acceptable tradeoffs amongst objectives. Together, these directions would help establish when world models can support reliable post-deployment adaptation. Conversely, if world-model predictions of policy performance after a change consistently fail to track realized performance, or if other approaches such as meta-learned or fault-randomized policies recover comparable performance under similar data and compute budgets, research effort should shift to those approaches. In summary, current world models have not yet demonstrated that they are reliable enough for deployed adaptation in unstructured environments for multiagent systems. However, given their potential, we argue that these limitations warrant further investigation into their use for multi-objective adaptation in remote robot teams rather than ruling them out.

\clearpage
\acknowledgments{The NASA Space Technology Graduate Research Opportunities 80NSSC26K1290 supported Jamie Santos' contributions to this paper. Thank you to Gaurav Dixit for supporting this work.}


\bibliography{example}  

\end{document}

%% file: glossary.tex
\newacronym{ml}{ML}{Machine Learning}
\newacronym{rl}{RL}{Reinforcement Learning}
\newacronym[
    plural={MDPs},
    longplural={Markov Decision Processes}
]{mdp}{MDP}{Markov Decision Process}
\newacronym[
    plural={POMDPs},
    longplural={Partially Observable Markov Decision Processes}
]{pomdp}{POMDP}{Partially Observable Markov Decision Process}
\newacronym[
    plural={Dec-POMDPs},
    longplural={Decentralized POMDPs}
]{decpomdp}{Dec-POMDP}{Decentralized Partially Observable MDP}
\newacronym{morl}{MORL}{Multi-Objective RL}
\newacronym{marl}{MARL}{Multiagent RL}
\newacronym{momarl}{MOMARL}{Multi-Objective Multiagent RL}
\newacronym{lta}{LTA}{Long-Term Autonomy}
\newacronym{auv}{AUV}{Autonomous Underwater Vehicle}
\newacronym{moma}{MOMA}{Multi-Objective Multiagent}
\newacronym{eas}{EAs}{Evolutionary Algorithms}
\newacronym{erl}{ERL}{Evolutionary Reinforcement Learning}
\newacronym{ctde}{CTDE}{Centralized Training with Decentralized Execution}
\newacronym{nsgaii}{NSGA-II}{Non-Dominated Sorting Genetic Algorithm II}
\newacronym{merl}{MERL}{Multiagent Evolutionary Reinforcement Learning}
\newacronym{mome}{MOME}{Multi-Objective MAP-Elites}
\newacronym{nsmerl}{NS-MERL}{Novelty Search MERL}
\newacronym[
    plural={PCNs},
    longplural={Pareto Conditioned Networks}
]{pcn}{PCN}{Pareto Conditioned Network}
\newacronym{hv}{HV}{hypervolume}
\newacronym{knn}{k-NN}{k-Nearest Neighbors}
\newacronym{maml}{MAML}{Multiagent Meta Learning}
\newacronym{difmaml}{Dif-MAML}{Decentralized Multiagent Meta Learning}
\newacronym{ood}{OOD}{Out of Distribution}
\newacronym{ma}{MA}{multiagent}
\newacronym{mo}{MO}{multi-objective}
\newacronym{pgmorl}{PG-MORL}{Prediction-Guided MORL}
\newacronym{qd}{QD}{Quality-Diversity}